\documentclass[10pt,twocolumn,english]{article}
\usepackage[T1]{fontenc}
\usepackage[latin9]{inputenc}
\usepackage{amsmath}
\usepackage{stmaryrd}
\usepackage{wasysym}

\makeatletter

\newcommand{\mathcircumflex}[0]{\mbox{\^{}}}

\usepackage{tikz-cd}
\usetikzlibrary{graphs} 
\tikzcdset{
arrow style = tikz,
diagrams={>={Straight Barb[scale=0.8]}}
}

\makeatother

\usepackage{babel}
\begin{document}

\title{General problem solving with category theory.}

\author{Francisco J. Arjonilla, Tetsuya Ogata}
\maketitle
\begin{abstract}
This paper proposes a formal cognitive framework for problem solving
based on category theory. We introduce cognitive categories, which
are categories with exactly one morphism between any two objects.
Objects in these categories are interpreted as states and morphisms
as transformations between states. Moreover, cognitive problems are
reduced to the specification of two objects in a cognitive category:
an outset (i.e. the current state of the system) and a goal (i.e.
the desired state). Cognitive systems transform the target system
by means of \emph{generators }and \emph{evaluators. }Generators realize
cognitive operations over a system by grouping morphisms, whilst evaluators
group objects as a way to generalize outsets and goals to partially
defined states. Meta-cognition emerges when the whole cognitive system
is self-referenced as sub-states in the cognitive category, whilst
learning must always be considered as a meta-cognitive process to
maintain consistency. Several examples grounded in basic AI methods
are provided as well.

\end{abstract}

\section{Introduction\protect\vphantom{Latex dependency workaround (do not remove)--> $\Circle\CIRCLE$<--}}

Unification of Artificial Intelligence (AI) has been a long pursued
goal since the early days of computation, however it remains elusive.
Here we propose a novel framework that lays the grounds of a formal
description of general intelligence. This framework is a generalization
and formalization of the concepts presented in \cite{Arjonilla2015},
which claimed that cognitive systems learn and solve problems by trial
and error: the attempts at reaching a goal, termed variants, are first
\emph{generated} by some heuristics and then assessed. The better
the models a cognitive system has, the more accurate the generation
of variants is and the fewer the mistakes made. Random variants are
inevitable when there are no models, and mistake-free variants are
are used when the model is complete.

Many theories have been proposed that attempt general problem solving,
yet the final goal of achieving human-level intelligence has been
unsuccessful. Some authors have proposed guidelines and roadmaps for
this search \cite{Rosa2016,Laird2010}. One of the earliest theories
that focus on general problem solving was proposed in \cite{Newell1959}
and focuses in decomposing recursively goals in subgoals and separating
problem content from problem solving strategies. It evolved later
into the cognitive architecture SOAR \cite{Laird2012,Laird1987} as
an example of a unified theory of cognition \cite{Newell1992}. Furthermore,
\cite{Hutter2000,Hutter2005b} proposed a general theory of universal
intelligence that combines Solomonoff induction with sequential decision
theory realized in a reinforcement learning agent called AIXI. However,
AIXI is incomputable \cite{Veness2011} and relies on approximations.
A formal measure of general intelligence was proposed in \cite{Legg2007}
and related it to AIXI. 

On the other hand, category theory has been seldom applied to modeling
general cognitive processes. Rather, these efforts have been directed
towards knowledge representation and specific cognitive processes.
\cite{Gomez-Ramirez2014} proposed a general framework for representation
based on category theory to advance in the understanding of brain
function. Other authors have focused in modeling the semantics of
cognitive neural systems \cite{Healy2000b}, describing certain aspects
of cognition such as systematicity \cite{Halford2010,Phillips2010},
or modeling theories about human consciousness such as Integrated
Information Theory \cite{Tsuchiya2016a,Tononi2004}.

\section{Category Theory}

Category theory is a relatively new field of mathematics and the theory
of structure \emph{par excelence}. It raises the importance of relations
between objects to that of the objects themselves. We now sketch the
categorical entities that stand as the formal skeleton of the cognitive
theory developed in this paper.

\textbf{\emph{Definition:}} A category $\mathcal{C}$ consists of
two entities:
\begin{enumerate}
\item A class $Obj(\mathcal{C})$ of elements. These elements are called
$objects$. An object $A\in Obj(\mathcal{C})$ is also written $A\in\mathcal{C}$.
\item Morphisms (Also maps or arrows): For each $A,B\in\mathcal{C},$ a
hom-set $hom_{\mathcal{C}}(A,B)$ whose elements $f\in hom_{\mathcal{C}}(A,B)$
are called the \emph{morphisms} from $A$ to $B$. $A$ is called
the \emph{domain }of $f$ and $B$ the \emph{codomain}. $f$ is also
written $f:A\longrightarrow B$ or $f_{AB}$. The class of all morphisms
in $\mathcal{C}$ is denoted as $Mor(\mathcal{C})$.
\end{enumerate}
With the following properties:
\begin{enumerate}
\item For each object $A\in\mathcal{C}$, there is a morphism $1_{A}\in hom_{\mathcal{C}}(A,A)$
called the identity morphism with the property that, for any morphism
$f\in hom_{\mathcal{C}}(A,B)$, then $1_{B}\circ f=f\circ1_{A}=f$.
\item Composition: If $f\in hom_{\mathcal{C}}(A,C)$ and $g\in hom_{\mathcal{C}}(C,B)$,
then there is a morphism $g\circ f\in hom_{\mathcal{C}}(A,B)$. $g\circ f$
is called the composition of $g$ with $f$.
\item Composition is associative: $f\circ(g\circ h)=(f\circ g)\circ h$.
\end{enumerate}
\hfill{}$\oblong$

There is no restriction on what elements can $\mathcal{C}$ hold.
In this article we will use small categories, i.e. $Obj(\mathcal{C})$
and $Mor(\mathcal{C})$ are sets. The elements of $\mathcal{C}$ may
be abstract mathematical entities, daily objects or even other categories.
In the latter case, the morphisms between categories receive a special
treatment and are called functors:

\textbf{\emph{Definition:}} A (covariant) functor $F:\mathcal{C}\longrightarrow\mathcal{D}$
is a morphism between categories \emph{$\mathcal{C},\mathcal{D}$}
with the following two components:
\begin{enumerate}
\item A function $F:Obj(\mathcal{C})\longrightarrow Obj(\mathcal{D})$ that
maps objects in $\mathcal{C}$ to objects in $\mathcal{D}$.
\item A function $F:Mor(\mathcal{C})\longrightarrow Mor(\mathcal{D})$ that
maps morphisms in $\mathcal{C}$ to morphisms in $\mathcal{D}$.
\end{enumerate}
With the following properties:
\begin{enumerate}
\item If $A,B\in\mathcal{C}$, $F$ maps a morphism $f:A\longrightarrow B$
to $Ff:FA\longrightarrow FB$.
\item Identity is preserved: $F1_{A}=1_{FA}$.
\item Composition is preserved: $F(g\circ f)=Fg\circ Ff$.
\end{enumerate}
\hfill{}$\oblong$

Other important concepts in category theory are natural transformations,
limits and adjunctions, which are not needed to introduce the concepts
presented here.

\section{Core}

Let us specify an \emph{independent system} as an entity that has
no relationship whatsoever with other systems. This definition grants
self-containment to systems that will undergo cognitive processing.
Independent systems are abstract, but will serve as an idealization
that will ease the study of non-independent systems. Moreover, we
assume that they abide at exactly one state in any given context.
We will also refer to cognitive systems. Cognitive systems will only
intervene on an independent system when the latter fails to fulfill
the desired goal. The other assumption is that the purpose of cognitive
systems is to solve problems specified as states that fulfill some
condition. We now present the key concepts along with some examples
grounded on AI.

\subsection{Cognitive categories}

Cognitive categories are the cornerstone of this framework. The generality
of category theory enables to apply this framework in a wide variety
of applications. We will first give the definition of cognitive category
and then study the properties.

\textbf{\emph{Definition:}} A cognitive category $\mathcal{S}$ is
a category that satisfies $|hom_{\mathcal{S}}(A,B)|=1$ for all $A,B\in\mathcal{S}$.\hfill{}$\oblong$

All objects in cognitive categories are initial and terminal objects.
Also, all morphisms are invertible and there are exactly $|Obj(\mathcal{S})|^{2}$
morphisms. Let us create an intuition for cognitive categories. The
objects $Obj(\mathcal{S})$ of a cognitive category $\mathcal{S}$
associated to an independent system, which we will also denote as
$\mathcal{S}$, correspond to each of the possible unique states that
the system can be found at. On the other hand, $hom_{\mathcal{S}}(A,B)$
with $A,B\in\mathcal{S}$ represents a transformation from state $A$
to state $B$. Other than that, we leave the specific mechanics of
transformations undefined. We will justify that this transformation
is unique as follows: It is always possible to conceive a transformation
between any two states by replacing the system in state $A$ with
another undistinguishable system in state $B$. For practical purposes,
transformation and replacement are equally effective because the relationships
between states of the system remain equal. Therefore $|hom_{\mathcal{S}}(A,B)|\geq1$.
Now, consider any $t_{1},t_{2}\in hom_{\mathcal{S}}(A,B)$. Since
$A$ and $B$ are states, the effects on the system of any transformation
$t_{1}\in hom_{\mathcal{S}}(A,B)$ will be undistinguishable from
any other transformation $t_{2}\in hom_{\mathcal{S}}(A,B)$, hence
$t_{1}=t_{2}$ and so, $|hom_{\mathcal{S}}(A,B)|=1$. Thus, the category
of states of an independent system is a cognitive category. With this
intuition in mind, identity morphisms are null transformations and
composition of $t_{1}:A\longrightarrow B$ and $t_{2}:B\longrightarrow C$
is a direct transformation $t_{21}:A\longrightarrow C$. Morphisms
in cognitive categories allow for the consideration of atomic transformations
between single states, but they are less useful when considering real
world problems because transformations are generally defined as operations
rather than a replacement of one specific state into another.

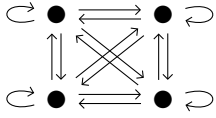
\begin{figure}
\begin{center} \nonumber \begin{tikzcd} 

& \CIRCLE 
\arrow[d,shift left=0.4ex]  \arrow[r,shift left=0.4ex]
\arrow[dr,shift left=0.4ex] \arrow[loop left=0.4ex] 
& \CIRCLE
\arrow[d,shift left=0.4ex]  \arrow[l,shift left=0.4ex]
\arrow[dl,shift left=0.4ex] \arrow[loop right=0.4ex]
\\

& \CIRCLE
\arrow[u,shift left=0.4ex]  \arrow[r,shift left=0.4ex]
\arrow[ur,shift left=0.4ex] \arrow[loop left=0.4ex] 
& \CIRCLE
\arrow[u,shift left=0.4ex]  \arrow[l,shift left=0.4ex]
\arrow[ul,shift left=0.4ex] \arrow[loop right=0.4ex]

\end{tikzcd} \end{center}

\caption{A cognitive category consisting of 4 states. The number of morphisms
is $4^{2}=16$.}
\end{figure}

\subsection{Cognitive problems\label{sub:CognitiveProblems}}

The next step is to characterize the functionality of a cognitive
system. We assumed that an independent system $\mathcal{S}$ may only
be found in one of its possible states at any one time. Consider that
this state is $O\in\mathcal{S}$ and consider another state $T\in\mathcal{S}$
that holds some desired condition. We will refer to $O$ as the \emph{outset
}and $T$ as the \emph{goal.}  The cognitive system receives these
states as fixed: $O$ is given by the current state of the independent
system and $T$ is given externally. The role of the cognitive system
is to find some method or operation that transforms the independent
system to a goal-complying state, and optionally executing this operation.
Once the cognitive system transforms the outset state into the goal
state, it becomes superfluous until another cognitive problem is presented. 

We can define for now a restricted definition of cognitive problem
that is valid when there is no uncertainty, which is covered in Section
\ref{sub:Evaluators} where $O$ and $T$ will be generalized to partially
known outsets and abstract goals by considering sets of outsets and
sets of goals.

\textbf{\emph{Definition: }}A \emph{deterministic cognitive problem}
$B$ is a triple $(\mathcal{S},O,T)$, where $\mathcal{S}$ is a cognitive
category, $O\in\mathcal{S}$ is the outset and $T\in\mathcal{S}$
the goal.\hfill{}$\oblong$

The solution to cognitive problems involves two stages, as mentioned
above:

First, find a morphism $t$ that guarantees that, if followed, the
solution will be reached. Since $|hom_{\mathcal{S}}(O,T)|=1$, $t$
is the singleton morphism in $hom_{\mathcal{S}}(O,T)$ and is fully
defined by $t\approx B=(\mathcal{S},O,T)$. However, the cognitive
system will not be able to provide $t$ reliably unless stored as
prior knowledge, which makes the base of model-based cognitions, as
we will see in section \ref{sub:Generators}.\begin{center} \nonumber \begin{tikzcd}

O \arrow[r, "?"] & T

\end{tikzcd} \end{center}

Second, follow $t$ to actually transform $\mathcal{S}$ from $O$
to $T$. Let us analyze further this stage. Assume that $\mathcal{S}$
is in state $O$ and that there is a morphism $t:O\longmapsto X$,
with X unknown, that is, the cognitive system knows the outset $O$
and knows that there is a morphism $t$ whose domain is $O$, but
has no knowledge about what state X will be reached if morphism $t$
is followed. The only way of finding $X$ is by following $t$. However,
there is an important drawback that originates from transforming $\mathcal{S}$:
The cognitive system might not have the ability to return $\mathcal{S}$
to $O$ after having followed $t$, that is, it might be uncapable
of solving $(\mathcal{S},T,O)$. Graphically,\begin{center} \nonumber \begin{tikzcd}

O \arrow[r, "t"] & ?

\end{tikzcd} \end{center}

The rest of the paper builds on how to find $t$ under various conditions
and shows through examples how this formulation is sufficient to describe
a variety of AI methods.

\textbf{\emph{Example:}} A Turing machine \cite{Turing1936} is a
formal model of computation that manipulates symbols in an infinite
string on a tape following a table of rules. Mathematically, it is
a 7-Tuple $M=\left\langle Q,\Gamma,b,\varSigma,\delta,q_{0},F\right\rangle $
\cite{Hopcroft1979} where the symbols denote, respectively, $Q:$
states (not to confuse them with states in independent systems), $\Gamma:$
tape alphabet symbols, $b:$ blank symbol, $\Sigma:$ input symbols,
$\delta:$ transition function, $q_{o}:$ initial state and $F:$
Accepting states. However, the control element of Turing machines
alone lack some characteristics of independent systems, specifically
that strings are not considered part of the state of the Turing machine.
Actually, the only discernible change between the initial and halting
conditions is that the internal state changes from $q_{0}$ to $q_{G}\in F$.
Hence the states of the independent system do not need to describe
the Turing machine. Moreover, we will assume that it always halts
and only consider the initial and halting contents of the tape. So,
we define the cognitive category of Turing machine strings $\mathcal{S}_{TM}$
whose objects are the set of all possible strings in the tape. The
morphisms replace one string with another, but hold no information
about what processes lie behind this transformation. That is the job
of the generators: A Turing machine is a component of a cognitive
system, specifically a generator, that operates on the tape and transforms
one string into another following the instructions set by its transition
function $\delta$. As purely computational machines, Turing machines
solve the second stage of cognitive problems as stated above.\hfill{}$\oblong$

One more comment: there is no single way of designing a cognitive
category. It depends on how a cognitive problem is best described.
For instance, we could have described instead the cognitive states
of a Turing machine in the previous example as the string after executing
each instruction. By way of composition of morphims between the initial
tape and the halting tape, this category can be reduced to the previous.

\subsection{Generators \label{sub:Generators}}

Morphisms in cognitive categories are well suited to study the atomic
transformations of independent systems, but they are most useful for
theoretical analyses. Generally, it is not possible to transform freely
a system. Rather we are constrained by a limited set of operations.
In practice, a cognitive system will perform operations on an independent
system regardless of its outset, with the resulting state depending
on the operation and the outset. Generators are cognitive solvers
that produces transformations for any state of $\mathcal{S}$ in the
scope of cognitive categories and stand as the first abstraction of
cognitive processes.

\textbf{\emph{Definition: }}A \emph{generator} $gt:Obj(\mathcal{S})\longrightarrow Obj(\mathcal{S})$
over a cognitive category $\mathcal{S}$ is an endomorphism in the
category of sets.\hfill{}$\oblong$

\begin{figure}
\begin{center} \nonumber \begin{tikzcd} 

& \CIRCLE 
\arrow[d,shift left=0.4ex,thick]  \arrow[r,shift left=0.4ex,dotted]
\arrow[dr,shift left=0.4ex] \arrow[loop left=0.4ex,dashed] 
& \CIRCLE
\arrow[d,shift left=0.4ex,dotted]  \arrow[l,shift left=0.4ex]
\arrow[dl,shift left=0.4ex,thick] \arrow[loop right=0.4ex,dashed]
\\

& \CIRCLE
\arrow[u,shift left=0.4ex,dotted]  \arrow[r,shift left=0.4ex,thick]
\arrow[ur,shift left=0.4ex] \arrow[loop left=0.4ex,dashed] 
& \CIRCLE
\arrow[u,shift left=0.4ex,thick]  \arrow[l,shift left=0.4ex,dotted]
\arrow[ul,shift left=0.4ex] \arrow[loop right=0.4ex,dashed]

\end{tikzcd} \end{center}

\caption{An omnipotent set of 4 generators over a cognitive category with 4
states. Dashed: Do not transform. Dotted: Cycle through all states.}
\end{figure}
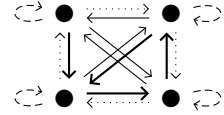

$gt$ stands for \emph{g}enerate and \emph{t}ransform. Generators
generalize transformations defined by a morphism to the whole set
of possible states. Intuitively speaking, a generator is the application
of an operation over $\mathcal{S}$ from an unspecific outset. More
specifically, it provides the morphism that is equivalent to running
the operation conveyed by the generator, and then follows the morphism.
Generators cover the first stage of problem solving as stated in section
\ref{sub:CognitiveProblems}. An isomorphic function is $g:Obj(\mathcal{S})\longmapsto hom_{\mathcal{S}}(O,T)$
such that $g$ is equivalent to $gt$ except that it does not follow
the morphism. If a state $X\in\mathcal{S}$, the relation between
$g$ and $gt$ is $gt(X)=\textrm{codom}(g(X))$ We will use $g$ or
$gt$ depending on which one fits best, knowing that they are isomorphic.

\begin{center} \nonumber \begin{tikzcd}
O \arrow[rr, "t_{O}=g(O)"] & & T=gt(O)
\end{tikzcd} \end{center}

The elements of $g$ are morphisms in $\mathcal{S}$ that have the
same properties as set functions (because they are set functions).
Specifically:
\begin{itemize}
\item $g$ provides one morphism for each state in $\mathcal{S}$.
\item $g$ is either an automorphism or both a non-injective and non-surjective
morphism in the category of sets: there are generators that cannot
transform $\mathcal{S}$ to every state.
\end{itemize}
\textbf{\emph{Example:}} Consider the cognitive category $\mathcal{S}_{M}$
of mathematical statements and a cognitive problem $B=(\mathcal{S}_{M},O\approx(x-1)^{2}=0,G\approx x=1)$.
The solution involves finding a function $\theta_{M}:(\mathcal{S},O,G)\longrightarrow g$
with $g:(x-1)^{2}=0\longmapsto x=1$. $\theta$ is well known for
$B$ (for example, \cite{CAS1992}) and computationally implemented
with Computer Algebra Systems. $\theta_{M}$ recognizes $O$ as a
polynomial, providing and calling a one-variable polynomial solver
for the generator $g$. Alternatively, a student $\theta_{H}$ provides
$g$ by writing a sequence of steps required to solve for $x$.\hfill{}$\oblong$

Consider the set $\mathcal{G}_{\mathcal{S}}$ of all possible generators
in $\mathcal{S}$. With respect to its cardinality, if $n=|\mathcal{S}|$
then for every object $A\in\mathcal{S}$ there are $n$ possible morphisms,
hence $|\mathcal{G}_{\mathcal{S}}|=n^{2}$. In general, cognitive
systems will not have access to $\mathcal{G}_{\mathcal{S}}$, but
rather to a subset of $\mathcal{G}_{\mathcal{S}}$ that constitute
the set of operations over $\mathcal{S}$ available to the cognitive
system. Let us study a special case of $G\in\mathcal{G}_{\mathcal{S}}$
that greatly simplifies the study of generators.

\textbf{\emph{Definition: }}An \emph{omnipotent} set of generators
over $\mathcal{S}$ is a set of generators $G\subset\mathcal{G}_{\mathcal{S}}$
that can generate any morphism in $\mathcal{S}$.\hfill{}$\oblong$

\textbf{\emph{Example:}} Consider an independent system with $n$
possible states associated to the cognitive category $\mathcal{S}=\{S_{1},\ldots,S_{n}\}$.
Consider as well a cognitive system with a unique generator $g$ over
$\mathcal{S}$ that cycles over all states: 
\[
g(S_{i})=\begin{cases}
S_{i+1} & \textrm{if }i\in\{1\ldots n-1\}\\
S_{1} & \textrm{if }i=n
\end{cases}
\]

An omnipotent set of generators is constructed by composing $g$ with
itself iteratively: $G=\{g_{i}=g\circ\overset{i}{\cdots}\circ g|i\in\{1\ldots n\}\}$\hfill{}$\oblong$

Given any outset O, a cognition with an omnipotent set of generators
can transform it to any other state and thus take complete control
of $\mathcal{S}$. 

\textbf{\emph{Theorem:}} Consider an omnipotent cognition over a cognitive
category $\mathcal{S}$ with a set of available generators $G\subset\mathcal{G}_{\mathcal{S}}$.
Then, $|G|\geq|\mathcal{S}|=n$.

\emph{Proof:} The number of morphisms that each generator can produce
is equal to $n=|\mathcal{S}|$. Taking the definition of cognitive
category, the number of morphisms in $\mathcal{S}$ is $n{}^{2}$.
Therefore, covering the full set of morphisms in $\mathcal{S}$ requires
a minimum of $n^{2}/n=n$ generators.\hfill{}$\oblong$

We will refer to G as a reduced set of generators when $|G|=n$. These
generators have interesting properties:
\begin{enumerate}
\item For each $t\in Mor(\mathcal{S})$ such that $t:A\longrightarrow B$
there exists a unique generator $g\in G$ such that $g:A\longmapsto B$.
\item For each $A\in\mathcal{S}$, and for each $g_{1},g_{2}\in G$, if
$g_{1}(A)=g_{2}(A)$ then $g_{1}=g_{2}$.
\end{enumerate}
The second property states that given an outset $O$, there exists
one generator for transforming to each state $T$. However, the opposite
is not true: The number of generators that transform each $O$ to
a fixed $T$ range from 0 to $n$. Generators are more interesting
when they transform any outset to a small number of states, which
leads to the next definition: 

\textbf{\emph{Definition:}} A \emph{purposeful }generator $g_{T}$
is the contravariant hom functor $\textrm{hom}(\cdot,T):\mathcal{S}\longmapsto\mathbf{set}$
defined by $g_{T}(O)=\textrm{hom}(\cdot,T)(O)=\textrm{hom}(O,T)=\{t_{OT}\}$.\hfill{}$\oblong$

We leave the analysis of the morphism component of this functor for
future works. The \emph{generate and transform }variety of $g$ may
also be used for an equivalent but simpler definition:

\textbf{\emph{Definition:}} A \emph{purposeful }generator $gt_{T}$
is the constant function $gt_{T}(O)=T$ where $O\in\mathcal{S}$.\hfill{}$\oblong$

The latter definition clearly shows that purposeful generators are
strongly related to states in the cognitive system. This is consistent
with the fact that the minimum number of generators required in an
omnipotent cognition matches the number of states in $\mathcal{S}$.
Actually, it is possible to build an omnipotent cognition with purposeful
generators exclusively from an omnipotent cognition with non-purposeful
generators: Consider a cognition with an initial omnipotent set of
non-purposeful generators $G$. Then, for every outset $O$ and every
goal $T$, there is a generator $g_{i}\in G$ such that $g_{i}(O)=t_{OT}$.
Now, consider a generator $g_{T}$ that uses the appropriate non-purposeful
generator for each possible outset, such that

\[
g_{T}(X)=\begin{cases}
g_{i}(X)=t_{O_{1}T} & \textrm{if }X=O_{1}\\
\vdots\\
g_{j}(X)=t_{O_{n}T} & \textrm{if }X=O_{n}
\end{cases}
\]

$g_{T}$ is indeed a purposeful generator because $gt_{T}(X)=T$ for
all $X\in\mathcal{S}$. We can repeat this argument for each state
in $\mathcal{S}$ and therefore build a set of $n$ distinct purposeful
generators:

\textbf{\emph{Definition:}} The \emph{canonical }set of generators
for the cognitive system $\mathcal{S}$ is the set $G_{\mathcal{S}}=\{g_{X}=hom(\cdot,X)|X\in\mathcal{S}\}$.\hfill{}$\oblong$

$G_{\mathcal{S}}$ is a reduced set of generators because $|G_{S}|=|Obj(\mathcal{S})|=n$.
Moreover, $G_{\mathcal{S}}$ is unique: Consider a reduced set of
purposeful generators $G'_{\mathcal{S}}$ that is built using a different
set of generators than $G_{\mathcal{S}}$. We have that for all $X,Y\in\mathcal{S}$,
there is a $g_{T}\in G_{\mathcal{S}}$ such that $g_{T}(X)=t_{XY}$
and a $g'_{T}\in G'_{\mathcal{S}}$ such that $g'_{T}(X)=t_{XY}$.
Taking into account that both sets have the same number of generators,
we conclude that $G_{\mathcal{S}}$ and $G'_{\mathcal{S}}$ are the
same set.

\begin{figure}
\begin{center} \nonumber \begin{tikzcd} 

& \CIRCLE 
\arrow[d,shift left=0.4ex,thick]  \arrow[r,shift left=0.4ex,dotted]
\arrow[dr,shift left=0.4ex] \arrow[loop left=0.4ex,dashed] 
& \CIRCLE
\arrow[d,shift left=0.4ex]  \arrow[l,shift left=0.4ex,dashed]
\arrow[dl,shift left=0.4ex,thick] \arrow[loop right=0.4ex,dotted]
\\

& \CIRCLE
\arrow[u,shift left=0.4ex,dashed]  \arrow[r,shift left=0.4ex]
\arrow[ur,shift left=0.4ex,dotted] \arrow[loop left=0.4ex,thick] 
& \CIRCLE
\arrow[u,shift left=0.4ex,dotted]  \arrow[l,shift left=0.4ex,thick]
\arrow[ul,shift left=0.4ex,dashed] \arrow[loop right=0.4ex]

\end{tikzcd} \end{center}

\caption{The canonical set of generators over a cognitive category with 4 states.
Morphisms with the same line style correspond to elements of the same
generator.}
\end{figure}
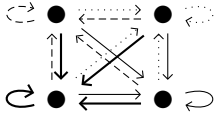

Morphisms in cognitive categories represent intuitively transformations
of states, however these transformations are too specific to be useful
in real problems. Normally a cognitive system will have a set of operations
to work with an independent system. The above developments show how
we can build operations that are specialized in pursuing a single
goal from a set of operations with unspecific results. More importantly,
we can assign one of these operations to each state such that instead
of specifying a goal as a state in $\mathcal{S}$, we can specify
a goal as an operation defined by a purposeful generator and extend
this result to the resolution of cognitive problems. Thus, any omnipotent
cognition over $\mathcal{S}$ straightforwardly solves any arbitrary
cognitive problem $B=(\mathcal{S},O,T)$ by calling the generator
$g_{T}\in G_{\mathcal{S}}$. Hence, at first sight $B$ could alternatively
be specified by the triple $(\mathcal{S},O,g_{T})$. 

\textbf{\emph{Example:}} Following the example above, the control
element of a Turing Machine $M$ is a generator over the infinite
string of symbols: For each state $O_{i}\in\mathcal{S}_{TM}$, $M$
produces another string $g_{TM}(O_{i})=T_{i}$. Turing machines that
overwrite the contents of the tape with a constant string are pseudo-purposeful
generators: $g_{TM}(O_{i})=T$ for all $O_{i}\in\mathcal{S}_{TM}$,
as long as $O_{i}$ and $T$ have a finite number of distinct elements.
Therefore, it is not possible to construct an omnipotent cognition
with Turing machines unless $Q$ is infinite, but by definition $Q$
is finite.\hfill{}$\oblong$

We have introduced the first steps towards formally grounding abstract
cognitive processing by showing how elementary transformations between
states are related to operations performed by cognitive systems on
independent systems. Generators generalize operations in the context
of category theory and provide a more natural way of approaching transformations
of independent systems than morphisms, which are limited to one specific
outset and goal each. On the contrary, the domain of generators include
every state in a cognitive category. Moreover, we have seen a class
of generators related to a single state each that take the target
system $\mathcal{S}$ to that state regardless of the outset and allows
to work with operations instead of states in cognitive problems.

\subsection{Evaluators\label{sub:Evaluators}}

Until now we have assumed that the states of $\mathcal{S}$ are fully
known. In general, this is not the case, however desirable. Evaluators
cover those situations where information about the state is needed,
yet there is partial or no access to this state. In a sense, evaluators
\emph{query} the independent system and return some evaluation on
the state, without actually referencing any particular state. The
power of evaluators stem from the fact that they can group many states
with a common property into a single object of another category, resulting
in an abstraction of unknown states into a cognitive category with
known states.

Consider a cognitive category $\mathcal{V}$ whose objects represent
abstract properties. The objects in $\mathcal{V}$ group states according
to some pattern, characteristic or common property. Also, consider
a functor $E:\mathcal{S}\Longrightarrow\mathcal{\mathcal{V}}$ that
sends each state in $\mathcal{S}$ to an object in $\mathcal{V}$.
$E$ establishes the relation between states in $\mathcal{S}$ and
abstract objects in $\mathcal{V}$.

\textbf{\emph{Definition:}} An evaluator is a functor $E:\mathcal{S}\Longrightarrow\mathcal{V}$,
with $\mathcal{S}$, $\mathcal{V}$ being cognitive categories.\hfill{}$\oblong$

\emph{$\mathcal{V}$ }is a partition of $\mathcal{S}$, therefore
the number of evaluators is precisely the Bell number $B_{n}$ \cite{Rota1964}
with index $n=|Obj(\mathcal{S})|$ :
\[
B_{n}=\frac{1}{e}\sum_{k=0}^{\infty}\frac{k^{n}}{k!}
\]

\begin{figure}
\begin{center} \nonumber \begin{tikzcd}

S_{A1} \arrow[d,"g(S_{A1})"'] & S_{A2} \arrow[l, "g(S_{A2})"']
\arrow[Rightarrow, rr, "ES_{A1}=ES_{A2}"] & & V_A 
\arrow[loop right, "g_\mathcal{V}(V_A)"]
\arrow[d,"g_\mathcal{V}(V_A)"]
\\

S_{B1} \arrow[r, "g(S_{B1})", shift left=0.4ex] &
S_{B2} \arrow[l, "g(S_{B2})", shift left=0.4ex]
\arrow[Rightarrow, rr, "ES_{B1}=ES_{B2}"] & & V_B
\arrow[loop right,"g_\mathcal{V}(V_B)"]

\end{tikzcd} \end{center}

\caption{Objects $S_{A1}$, $S_{A2}$, $S_{B1}$ and $S_{B2}$ belong to the
cognitive category $\mathcal{S}$, and $V_{A},V_{B}\in\mathcal{V}$.
Only the morphisms generated by $g$ and its counterpart $g_{\mathcal{V}}$
are shown. Evaluator $E:\mathcal{S}\protect\longrightarrow\mathcal{V}$
sends objects $S_{A1}$ and $S_{A2}$ to $V_{A}$, and $S_{B1}$ and
$S_{B2}$ to $V_{B}$. Generator $g$ cannot be transferred to \emph{$\mathcal{V}$
}because its $g_{\mathcal{V}}$ is not a function.}
\end{figure}

\textbf{\emph{Example:}} Neural networks for classification \cite{Zhang2000}
take input vectors and output an element in a finite set. It is possible
to construct an evaluator from these neural networks. If \emph{$\mathcal{S}$}
is a cognitive category whose states $\mathbf{s}_{i}$ are described
with vectors and $\mathcal{V}$ is another cognitive category whose
objects $V\in\mathcal{V}$ are the output elements of a neural network
$N:\mathbf{s}_{i}\longrightarrow Obj(\mathcal{V})$, then an evaluator
$E:\mathcal{S}\longrightarrow\mathcal{V}$ is constructed in the following
way: taking $\mathbf{s}\in\mathcal{S}$, the object component of $E$
is the neural network: $E(\mathbf{s})=N(\mathbf{s})$ and the morphism
component is trivial: $E(t_{\mathcal{\mathbf{rs}}})=t_{E(\mathbf{r})E(\mathbf{s})}$.\hfill{}$\oblong$

\textbf{\emph{Example:}} Consider a cognitive problem where the desired
goal is a positive assessment from all binary evaluators\emph{ $E{}_{1}\cap\ldots\cap E{}_{n}$},
i.e. with two outcomes \emph{pass }and \emph{fail}. In this case,
the appropriate evaluator to use as goal for the cognitive problem
is $E(S)=E_{1}(S)\mathcircumflex\ldots\mathcircumflex E_{n}(S)$.\hfill{}$\oblong$

\subsubsection{Generators in evaluators.}

Evaluators are only useful if we can apply transformations to their
outcomes. Let us transfer the generators over $\mathcal{S}$ to generators
over $\mathcal{V}$. $\mathcal{V}$ is an abstract category, so it
does not make sense to talk about operations in $\mathcal{V}$ unless
they are grounded somehow. For that reason, the operations conveyed
by generators $g'_{i}$ over $\mathcal{V}$ are the same operations
as generators $g_{i}$ over $\mathcal{S}$, hence $gt'$ (generate
and transform) actually transforms the outset in \emph{$\mathcal{S}$.}
If $E$ is an evaluator, we say that $g'=Eg$ and take that $g'$
is the set of codomains of $E$ restricted to the morphisms generable
by $g$.

\textbf{\emph{Definition:}} Given a set of generators $G_{\mathcal{S}}$
over $\mathcal{S}$, an\emph{ }evaluator $E:\mathcal{S}\Longrightarrow\mathcal{V}$
is \emph{controllable }by $G_{\mathcal{S}}$ if $G'_{\mathcal{S}}=EG_{\mathcal{S}}=\{g_{i}'=Eg_{i}|g_{i}\in G_{\mathcal{S}}\}$
forms an omnipotent set of generators over $\mathcal{V}$.\hfill{}$\oblong$

Cognitive problems expressed with controllable evaluators allow for
complete control over a system while allowing uncertainty in the definition
of the outset and goal.

\textbf{\emph{Example:}} John loves Mary, but he does not know if
she loves him back. John evaluates all possible states in the universe
$\mathcal{U}$ in 2 outcomes: Those where Mary loves John $\heartsuit$
and those that do not $\not\heartsuit$. His love is so deep that
he does not even consider an universe where either himself or Mary
do not exist. His cognitive problem is $(\mathcal{U},\not\heartsuit\cup\heartsuit=Obj(\mathcal{U}),\heartsuit)$.
John decides to take action $g=$``confess his love in public''
to make her fall in love with him, unaware that $g$ is not a purposeful
generator because Mary will feel embarrassed if she loved him. John
loses the love of her life.

\begin{center} \nonumber \begin{tikzcd}

\heartsuit \arrow[rr, "\textrm{confess}(\heartsuit)",shift left=0.4ex] & &
\not \heartsuit \arrow[ll, "\textrm{confess}(\not \heartsuit)",shift left=0.4ex] 

\end{tikzcd} \end{center}

This example shows that it is possible to model abstract concepts
as cognitive categories with the use of generators and evaluators,
without requiring extensive knowledge of the underlying independent
system.\hfill{}$\oblong$

\subsubsection{Hidden states.}

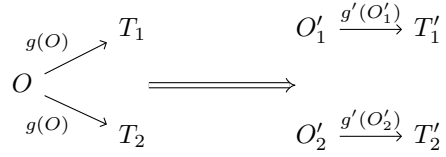
\begin{figure}
\begin{center} \nonumber \begin{tikzcd}[row sep=tiny]

& T_1 & &
O'_1 \ar[r, "g'(O'_1)"] & T'_1 \\

O  \ar[ur, "g(O)"] \ar[dr, "g(O)"']
& \text{ } \ar[rr,Rightarrow] & & \text{ } & \\
& T_2 & & O'_2 \ar[r, "g'(O'_2)"] & T'_2

\end{tikzcd} \end{center}

\caption{\label{fig:HiddenStates} Outset $O$ presents one hidden state because
$g(O)$ is not a function. $O$ is splitted into $O'_{1}$ and $O_{2}'$
and $g$ adapted to function $g'$.}
\end{figure}

By definition, a generator assigns one morphism (or state) to each
state. However, due to uncertainties, some operations do not always
led to the same result, even if that operation is repeated from the
same outset. Therefore, the operation does not yield a generator because
it is not a function. The solution is to consider hidden states (see
Figure \ref{fig:HiddenStates}). 

Consider an outset $O\in\mathcal{S}$ and a putative generator $g\in\mathcal{G}_{\mathcal{S}}$
that such that 
\[
g(O)=\begin{cases}
T_{1} & P(g(O)=T_{1}|O)=p_{1}\\
T_{2} & P(g(O)=T_{2}|O)=p_{2}=1-p_{1}
\end{cases}
\]

where $p_{1}$ and $p_{2}$ are the probabilities that the outcome
of $g(O)$ are $T_{1}$ and $T_{2}$, respectively. $g$ is not a
function so it is not a generator. This is the trick: We split the
outset $O$ into \emph{$O'_{1}$ }and $O'_{2}$ in another cognitive
category $\mathcal{S}'$ such that the new category has the same states
as $\mathcal{S}$ plus an additional one that represents the second
outcome from $g(O)$, and construct a proper generator\emph{ $g'$
}such that if $X\neq O$ and $g(X)=Y$, then $g'(X')=Y'$, and if
$X=O,$ then $g'(O'_{1})=T'_{1}$ and $g'(O'_{2})=T'_{2}$. That is,
$O$ is equivalent to $O_{1}$ when $g(O)=T_{1}$ and analogously
to $O_{2}$. We now have grounded the generator in a more accurate
cognitive category and removed the uncertainty posed by g(O). By application
of an evaluator $E:\mathcal{H}\Longrightarrow\mathcal{S}$ 
\[
E(X')=\begin{cases}
O & \textrm{if }X'=O'_{1}\textrm{ or }X'=O'_{2}\\
X & \textrm{Otherwise}
\end{cases}
\]
 we can verify that $Eg'$ is not a generator in $\mathcal{S}$. $O'_{1}$
and $O'_{2}$ are hidden states in $O$.

The bottom line is that an operation over a cognitive category $\mathcal{S}$
that yields two different outcomes from the same outset is an indication
that there is some deeper structure that we are not aware of, yet
it is still possible to be modeled as a cognitive category by designing
a more detailed model that takes the different outcomes as a substate
of the outset. Hidden states are reciprocal to evaluators: Evaluators
reduce the number of states in a category whilst hidden states increments
them.

We are now ready to define cognitive problems in a general form.

\textbf{\emph{Definition: }}Given a cognitive category $\mathcal{S}$,
a\emph{ cognitive problem} $B$ is a triple $(\mathcal{S},\mathbf{O},\mathbf{T})$,
with $\mathbf{O},\mathbf{T}\subset\mathcal{S}$ subsets denoting outset
and goal, respectively.\hfill{}$\oblong$

This definition has the advantage of expressing a cognitive problem
with partial information about states of $\mathcal{S}$. An equivalent
formulation is to define the outsets and goals implicitly as objects
$V_{O},V_{T}\in\mathcal{V}$ in the codomain of an evaluator $E:\mathcal{S}\longrightarrow\mathcal{V}$.
The solution to $B$ is a generator that sends every object $O_{i}\in\mathbf{O}$
to any object $T_{j}\in\mathbf{T}$. This way, the actual outset,
whose uncertainty is represented by the set $\mathbf{O}$, is guaranteed
to be sent to any state $T_{j}$ that complies with the desired goal.

\textbf{\emph{}}

\textbf{\emph{Example: }}In this example we will model genetic algorithms
with the proposed framework. Genetic algorithms (see for example \cite{Whitley1994}
or \cite{Goldberg2006}) are iterative processes that are composed
of (1) a population of $n$ \emph{chromosomes }$\mathbf{v}_{i}$ generally
represented with bit vectors, (2) genetic operators that alter the
chromosomes, and (3) a fitness function $f$. The adaptation is as
follows: (1) The population of chromosomes $\mathbf{v}^{n}$ maps
to one object in the cognitive category $\mathcal{P}$ of populations,
(2) genetic operators map to generators over $\mathcal{P}$, and (3)
fitness functions map to evaluator \emph{$E:\mathcal{P}\longrightarrow\mathcal{V}$},
where $Obj(\mathcal{V})=\{V_{opt},V_{not}\}$ representing, respectively,
better optimization than the outset population and not better\emph{.}
However, mutation and crossover operators yield random morphisms in
$\mathcal{P}$, which indicate the presence of hidden states that
encode the future outcomes of the mutation operators, i.e. the seed
in pseudo-random generators. To overcome this uncertainty, we pose
the cognitive problem by specifying outset and goal in $\mathcal{V}$:
$B_{opt}=(\mathcal{\mathcal{P}},V_{not},V_{opt})$. Now, mutation
and crossover operations have a high probability of being proper generators
in $\mathcal{V}$. Note that we are not constructing proper generators
from pseudo-generators, but rather bypassing the randomness presented
in $\mathcal{P}$ by implicitly considering the cognitive category
of hidden states $\mathcal{H}_{\mathcal{P}}$, an evaluator $F:\mathcal{H}\longrightarrow\mathcal{P}$
and the composition $E\circ F:\mathcal{H}\longrightarrow\mathcal{V}$,
which allows to send generators from $\mathcal{H}$ to \emph{$\mathcal{V}$}
if the parameters of the genetic algorithm are chosen adequately to
converge.\begin{center} \nonumber \begin{tikzcd}

\mathcal{H} \ar[r,"F"] \ar[rr,"E\circ F",bend right]
& \mathcal{P} \ar[r,"E"]
& \mathcal{V}

\end{tikzcd} \end{center}\hfill{}$\oblong$

\subsection{Dynamic systems}

We have previously stated that independent systems keep no relation
to other systems whatsoever. This also applies to time. In order to
conserve generality, we give time no special consideration and incorporate
it into states, which yields the expected independency. If $\mathcal{S}_{D}$
is a dynamic independent system, let us describe a state $S\in\mathcal{S}_{D}$
as a vector of static substates $S=\{S_{t_{0}},S_{t_{1}},\ldots\}$.
Thus, $S$ holds the complete timeline of $\mathcal{S}_{D}$ in a
single state. The set of objects in the cognitive category $\mathcal{S}_{D}$
holds all the possible timelines. We deal with uncertainty in time
by considering deterministic (causal determinism) and indeterministic
dynamic systems.

\subsubsection{Time in deterministic systems.}

Consider a dynamic independent system whose state is defined by $S=\{S_{t_{0}},S_{t_{1}},\ldots\}$.
A deterministic independent system is completely determined by the
state in one instant $t_{0}$. The remaining instants are calculated
inductively from $S_{t+1}=f(S_{t})$. Hence, we can rewrite the state
of a deterministic system with just $S=S_{0}$ and use it for the
objects in the cognitive category.

\subsubsection{Time in indeterministic systems.}

In this case, the states $S\in\mathcal{S}$ must include the temporal
evolution of the system in all instants to completely and uniquely
describe each possible timeline, i.e. two timelines that split at
instant $t_{1}$ are described by states $S_{A}=\{S_{t_{0}},S_{t_{1}}^{A},\ldots\}$
and $S_{A}=\{S_{t_{0}},S_{t_{1}}^{B},\ldots\}$. If a cognition has
access to evaluating $\mathcal{S}$ only at instant $t_{0}$, then
we can construct an auxiliary cognitive category $\mathcal{V}$ whose
objects are all the possible states of $\mathcal{S}$ only at instant
$t_{0}$: $\mathcal{V}=\{V_{i}=S_{t_{0}}^{i}\}$ and an evaluator
$E:\mathcal{S}\longrightarrow\mathcal{V}$ such that $E(S_{i})=S_{t_{0}}^{i}$.
The result is that we have a cognitive category, i.e. $\mathcal{V}$,
of an indeterministic dynamical system with uncertainty in future
instants, but subject to cognitive processing with all the formal
tools of the previous sections.

\subsection{Agents}

The agent-environment paradigm is prevalent in artificial intelligence.
In this paper we drop the assumption that agents and environments
should be treated as separated components with the definition of independent
systems. This proposal integrates the agent-environment paradigm as
a special case of independent systems. To start with, consider an
independent system $\mathcal{S}$ consisting of two subsystems Agent
$\mathcal{S}_{A}$ and Environment $\mathcal{S}_{E}$. Then all states
$S\in\mathcal{S}$ have the form $S=A\times E$, where $A\in\mathcal{S}_{A}$
and $B\in\mathcal{S}_{B}$ are substates that describe the agent and
the environment, respectively. A putative cognitive system acts on
$\mathcal{S}$ as a whole, rather than controlling agent and/or environment
separately.

\textbf{\emph{}}

\textbf{\emph{Example:}} A neural network that controls an automatic
car is described by a set of weights $\bar{w}=\{w_{i,j}\}$. The timeline
of the neural network, car and roadways constitute an independent
system $\mathcal{S}=A\times E$ where $A=\bar{w}\times A_{C}\times A_{M}$
describes the neural network $\bar{w}$, the computing system $A_{C}$
that executes the neural network, and the mechanical components that
drive the car and sense the environment $A_{M}$, whilst E describes
the roadway. $A$ drives safely along $E$ with no need for a cognitive
system. Even so, we would like to improve the reliability of the system
against unexpected situations, so we take a deep learning algorithm
to train the neural network for better performance. This algorithm
executes independently to driving the car and transforms $\mathcal{S}$
by modifying the values in $\bar{w}$, therefore the deep learning
algorithm is a generator over the category of neural networks that
drive $A$, and it solves the cognitive problem $B=(\mathcal{S},\mathbf{O},\mathbf{T})$,
where $\mathbf{O}=\bar{w}\times\mathbf{A}_{C}\times\mathbf{A}_{M}\times\mathbf{E}\subset\mathcal{S}$
is all the possible states that describe the independent system with
a fixed $\bar{w}$ and $\mathbf{T}\subset\mathcal{S}$ is the set
of all states where the neural network performs better than any $O\subset\mathbf{O}$.
Moreover, consider the evaluator $F:\bar{w}\times A_{C}\times A_{M}\times E\longrightarrow\bar{w}$,
or equally, $F:\mathcal{S}\longrightarrow\mathcal{V}_{\bar{w}}$.
Any generator over $\mathcal{S}$ will only be an omnipotent generator
over at most $\mathcal{V}_{\bar{w}}$ because there are no generators
that transform $A_{C}$, $A_{M}$ or $E$ independently to $\bar{w}$.\hfill{}$\oblong$

\section{Discussion}

We have succintly introduced the principles of a new way to understand
artificial intelligence. There is a long path until we can fully understand
how category theory may contribute to cognitive theories. Indeed,
just from applying two of the most basic concepts in category theory,
i.e. categories and functors, we have shown that turing machines,
neural networks and evolutionary algorithms admit a single formalization
under the proposed framework. This framework considers operations
as \emph{black boxes} and constructs methods to manage them and operate
with them. The flexibility of categories allows to consider many different
kinds of systems which are not limited to the target system, but can
also represent abstract systems grounded on the target system that
ease the theoretical study of the operations.

It remains to study how other central concepts in category theory
can contribute to extend this framework further, for example by analyzing
the role of the Yoneda Lemma in deepening our understanding of canonical
sets of generators (section \ref{sub:Generators}). Moreover, one
of the central cognitive abilities that has been left out of this
article is analogies. It seems plausible to generalize evaluators
and hidden states to analogies using adjunctions instead of functors,
which would enable to construct a network of adjunctions between cognitive
categories of representations. Furthermore, we postulate that cognitive
categories of generators and of evaluators enable a cognitive system
to cognitively process and improve its own methods, opening the path
to formally describing meta-cognition by fusing the cognitive system
with the independent system it controls. We leave these topics for
future developments, as well as more detailed cognitive-categorical
models of the examples presented. 

We have challenged some prior assumptions traditionally rooted in
AI. Firstly, we have dropped the distinction between agent and environment
in order to process them as a whole. The advantage is that any behaviour
that emerges from interactions agent-environment requires no special
treatment. Secondly, we considered time as an additional state variable
by including the timeline of the system as states. For this reason,
we claim that the study of general AI demands that we approach it
from several perspectives, challenging the assumptions that may hinder
the progress in the field, to unblock the switch from narrow to general
AI.

\bibliographystyle{plain}
\bibliography{\string"/Users/paco/Documents/BibTeX/80 - 2017 Arjonilla-805 - Insertados\string"}

\begin{thebibliography}{10}

\bibitem{Arjonilla2015}
Francisco~J Arjonilla.
\newblock {\em {A three-component cognitive theory}}.
\newblock Msc. thesis, Universiteit Utrecht, 2015.

\bibitem{CAS1992}
Keith~O. Geddes, Stephen~R. Czapor, and George Labahn.
\newblock {\em {Algorithms for Computer Algebra}}.
\newblock Springer, 1992.

\bibitem{Goldberg2006}
David~E. Goldberg.
\newblock {\em {Genetic Algorithms}}.
\newblock Pearson Education India, 2006.

\bibitem{Gomez-Ramirez2014}
Jaime G{\'{o}}mez-Ram{\'{i}}rez.
\newblock {\em {A New Foundation for Representation in Cognitive and Brain
  Science}}.
\newblock Springer Netherlands, Dordrecht, 2014.

\bibitem{Halford2010}
Graeme~S. Halford, William~H. Wilson, and Steven Phillips.
\newblock {Relational knowledge: the foundation of higher cognition.}
\newblock {\em Trends in Cognitive Sciences}, 14(11):497--505, nov 2010.

\bibitem{Healy2000b}
M.J. Healy.
\newblock {Category theory applied to neural modeling and graphical
  representations}.
\newblock {\em Proceedings of the IEEE-INNS-ENNS International Joint Conference
  on Neural Networks. IJCNN 2000. Neural Computing: New Challenges and
  Perspectives for the New Millennium}, pages 35--40 vol.3, 2000.

\bibitem{Hopcroft1979}
John~E. Hopcroft and Jeffrey~D. Ullman.
\newblock {\em {Introduction to Automata Theory, Languages and Computation}}.
\newblock Addison-Wesley Publishing Company, 1979.

\bibitem{Hutter2000}
Marcus Hutter.
\newblock {A Theory of Universal Artificial Intelligence based on Algorithmic
  Complexity}.
\newblock 2000.

\bibitem{Hutter2005b}
Marcus Hutter.
\newblock {\em {Universal Artificial Intelligence}}, volume~1 of {\em Texts in
  Theoretical Computer Science An EATCS Series}.
\newblock Springer Berlin Heidelberg, Berlin, Heidelberg, 2005.

\bibitem{Laird2012}
John~E. Laird.
\newblock {\em {The Soar Cognitive Architecture}}.
\newblock 2012.

\bibitem{Laird1987}
John~E Laird, Allen Newell, and Paul~S Rosenbloom.
\newblock {SOAR: An integrative architecture for general intelligence}.
\newblock {\em Artificial Intelligence}, 33:1--64, 1987.

\bibitem{Laird2010}
John~E. Laird and Robert~E. Wray.
\newblock {Cognitive architecture requirements for achieving AGI}.
\newblock In {\em Proc. of the Third Conference on Artificial General
  Intelligence}, pages 79--84, 2010.

\bibitem{Legg2007}
Shane Legg and Marcus Hutter.
\newblock {Universal Intelligence: A Definition of Machine Intelligence}.
\newblock {\em Minds and Machines}, 17(4):391--444, dec 2007.

\bibitem{Newell1959}
A.~Newell, J.~C. Shaw, and H.~A. Simon.
\newblock {Report on a general problem-solving program}.
\newblock In {\em IFIP Congress}, volume 224, 1959.

\bibitem{Newell1992}
Allen Newell.
\newblock {Unified theories of cognition and the role of Soar}.
\newblock In {\em Soar A cognitive architecture in perspective A tribute to
  Allen Newell Studies in cognitive systems Vol 10}, pages 25--79 ST -- Unified
  theories of cognition and the. 1992.

\bibitem{Phillips2010}
Steven Phillips and William~H. Wilson.
\newblock {Categorial compositionality: A category theory explanation for the
  systematicity of human cognition}.
\newblock {\em PLoS Computational Biology}, 6(7):7, 2010.

\bibitem{Rosa2016}
Marek Rosa, Jan Feyereisl, and {The GoodAI Collective}.
\newblock {A Framework for Searching for General Artificial Intelligence}.
\newblock 2016.

\bibitem{Rota1964}
Gian-Carlo Rota.
\newblock {The Number of Partitions of a Set}.
\newblock {\em The American Mathematical Monthly}, 71(5):498--504, 1964.

\bibitem{Tononi2004}
Giulio Tononi.
\newblock {An information integration theory of consciousness.}
\newblock {\em BMC neuroscience}, 5:42, nov 2004.

\bibitem{Tsuchiya2016a}
Naotsugu Tsuchiya, Shigeru Taguchi, and Hayato Saigo.
\newblock {Using category theory to assess the relationship between
  consciousness and integrated information theory}.
\newblock {\em Neuroscience Research}, 107:1--7, 2016.

\bibitem{Turing1936}
Alan Turing.
\newblock {On computable numbers}.
\newblock {\em Proceedings of the London Mathematical Society}, 25(2):181--203,
  1936.

\bibitem{Veness2011}
Joel Veness, Kee~Siong Ng, Marcus Hutter, William Uther, and David Silver.
\newblock {A Monte-Carlo AIXI approximation}.
\newblock {\em Journal of Artificial Intelligence Research}, 40:95--142, 2011.

\bibitem{Whitley1994}
Darrell Whitley.
\newblock {A genetic algorithm tutorial}.
\newblock {\em Statistics and Computing}, 4(2), jun 1994.

\bibitem{Zhang2000}
G.P. Zhang.
\newblock {Neural networks for classification: a survey}.
\newblock {\em IEEE Transactions on Systems, Man and Cybernetics, Part C
  (Applications and Reviews)}, 30(4):451--462, 2000.

\end{thebibliography}

\end{document}